\documentclass[letterpaper]{article} % DO NOT CHANGE THIS
\usepackage[preprint]{aaai2027}  % DO NOT CHANGE THIS
\usepackage[hyphens]{url}  % DO NOT CHANGE THIS
\usepackage{graphicx} % DO NOT CHANGE THIS
\usepackage{natbib}  % DO NOT CHANGE THIS AND DO NOT ADD ANY OPTIONS TO IT
\usepackage{caption} % DO NOT CHANGE THIS AND DO NOT ADD ANY OPTIONS TO IT
\usepackage{algorithm}
\usepackage{algorithmic}

\usepackage{newfloat}
\usepackage{listings}
\DeclareCaptionStyle{ruled}{labelfont=normalfont,labelsep=colon,strut=off} % DO NOT CHANGE THIS
\floatstyle{ruled}
\newfloat{listing}{tb}{lst}{}
\floatname{listing}{Listing}

\usepackage{booktabs}

\usepackage{amssymb}
\usepackage{amsmath}
\newtheorem{theorem}{Theorem}

\title{MAEM: Mitigating Recency Bias via a Replay-Free, Capacity-Controlled Dynamic Mixture of Experts for Class-Incremental Graph Learning}
\author{
Junlin Chen\textsuperscript{\rm 1},
Wenhao Wang\textsuperscript{\rm 1},
Junchao Xu\textsuperscript{\rm 1},
Qingsong Yao\textsuperscript{\rm 2},
Fei Xue\textsuperscript{\rm 3},\\
Li Sun\textsuperscript{\rm 4},
Xiao Wang\textsuperscript{\rm 1},
Ruijie Wang\textsuperscript{\rm 1}
}
\affiliations{
\textsuperscript{\rm 1}Beihang University, Beijing, China\\
\textsuperscript{\rm 2}Renmin University of China, Beijing, China\\
\textsuperscript{\rm 3}Trine University, Indiana, USA\\
\textsuperscript{\rm 4}Beijing University of Posts and Telecommunications, Beijing, China\\
}

\begin{document}

\maketitle

\begin{abstract}
Graph Neural Networks (GNNs) excel on graph-structured data but suffer catastrophic forgetting when classes arrive sequentially. In class-incremental graph learning, predictions are often dominated by recently learned classes, causing severe recency bias. Dynamic expansion can isolate knowledge across sessions, but many methods assume task-incremental settings and require task identities at test time to select branches or heads---information unavailable in class-incremental learning. Learnable routers offer an alternative, yet continual fine-tuning the router on new data can bias routing toward recent components.
We propose MAEM, a replay-free, capacity-controlled dynamic mixture-of-experts framework for class-incremental graph learning. MAEM builds representations with Simplified Graph Convolution (SGC) and trains a dedicated expert for each new session, then freezes it to protect prior knowledge. For routing without task IDs or a shared trainable router, each expert uses a masked autoencoder (MAE); test nodes are assigned to the expert with the lowest reconstruction error, reducing routing-level recency bias. When the expert pool reaches a fixed capacity, MAEM merges the least-used experts via Git Re-Basin alignment and further distills them into a fused expert using topology-aware pseudo data synthesized from MAE distributions (DeepInversion-style), without storing or revisiting historical graph data. On four class-incremental graph benchmarks, MAEM mitigates recency bias, outperforms strong continual graph learning baselines, and keeps model capacity bounded under strict replay-free learning.
\end{abstract}

% Uncomment the following to link to your code, datasets, an extended version or similar.
% You must keep this block between (not within) the abstract and the main body of the paper.
% Make sure that you do not de-anonymize yourself with these links.
% \begin{links}
%     \link{Code}{https://aaai.org/example/code}
%     \link{Datasets}{https://aaai.org/example/datasets}
%     \link{Extended version}{https://aaai.org/example/extended-version}
% \end{links}

\section{Introduction}

Graph Neural Networks (GNNs) have achieved strong performance in learning from graph-structured data \cite{kipf2017gcn,hamilton2017graphsage,velickovic2018gat}. However, real-world graphs often evolve continuously, with new nodes and classes arriving over time \cite{zhang2022cglb}. Class-incremental graph learning requires a model to learn newly emerging classes while retaining its ability to recognize all previously observed ones \cite{vandeven2019three,zhang2022cglb}. In this setting, catastrophic forgetting is primarily manifested as \emph{recency bias}: because only the newly introduced classes are supervised during the current session, the continually updated model becomes disproportionately biased toward these classes \cite{hou2019unified,wu2019large}. Consequently, nodes from earlier sessions are increasingly misclassified as belonging to the most recent classes.
\begin{figure}[t]
    \centering
    \includegraphics[width=0.6\columnwidth]{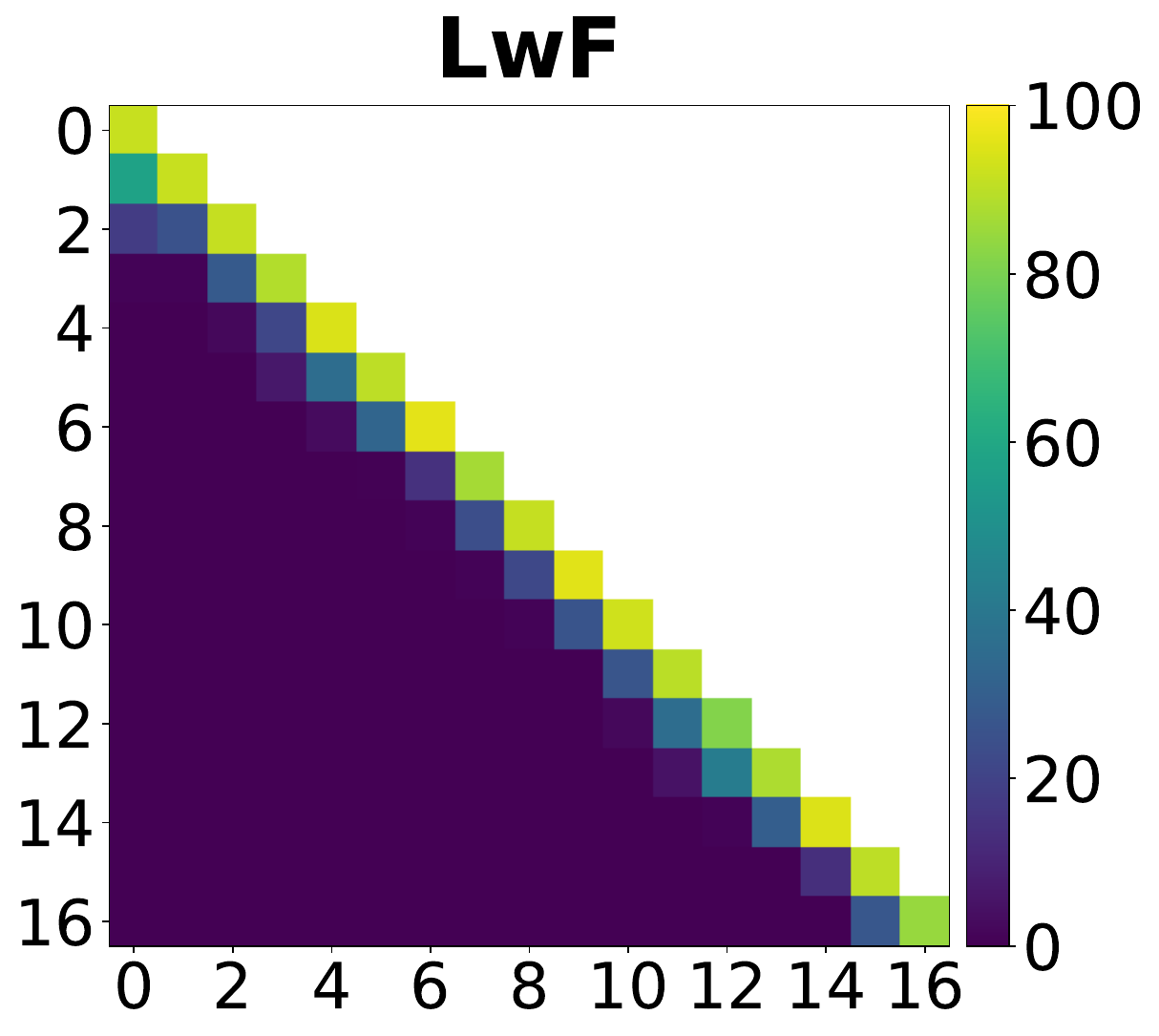}
    \caption{Session-wise accuracy matrix of LwF on cora-full. The entry in row $i$ and column $j$ denotes the accuracy on session $j$ after training through session $i$.}
    \label{fig:lwf_heatmap}
\end{figure}
Figure~\ref{fig:lwf_heatmap} illustrates this problem using LwF on cora-full \cite{bojchevski2018deep}. Although LwF preserves previous predictions through knowledge distillation, its accuracy on earlier sessions rapidly decreases as new classes are learned \cite{li2017lwf}. 
Existing continual learning methods mainly rely on regularization, replay, or architecture expansion \cite{parisi2019continual,delange2022continual}. Regularization-based methods constrain model updates but still repeatedly optimize a shared model on newly arriving classes, which cannot effectively suppress recency bias since such constraints preserve previously learned knowledge only approximately \cite{kirkpatrick2017ewc,aljundi2018mas,liu2021twp}. Replay-based methods recover historical knowledge by storing previous samples, which introduces additional memory costs and may violate data-retention constraints \cite{rebuffi2017icarl,lopezpaz2017gem,zhou2021ergnn}. In contrast, architecture-expansion methods allocate separate components to different sessions, naturally reducing interference between old and new knowledge \cite{rusu2016pnn,yoon2018den,aljundi2017expertgate}.
Nevertheless, existing expansion-based methods face two important limitations. First, many of them are designed for task-incremental learning and require the task identity of each test instance to select the corresponding branch or prediction head \cite{vandeven2019three,rusu2016pnn}. Progressive Neural Networks (PNNs), for example, assign a separate column to each task and normally rely on task identity during inference \cite{rusu2016pnn}. This information is unavailable in class-incremental learning, where a test node must be classified among all classes observed so far \cite{vandeven2019three}. Second, replacing task identities with a learnable router does not fully solve the problem. Because the router is continually fine-tuned on newly arriving data, it can itself develop recency bias and preferentially assign nodes to recently added experts. Moreover, continuously creating new components leads to unbounded model growth \cite{rusu2016pnn,yoon2018den}.
To address these limitations, we propose \textbf{MAEM}, a replay-free and capacity-controlled dynamic Mixture-of-Experts framework for class-incremental graph learning. MAEM first employs Simplified Graph Convolution (SGC) to obtain graph representations \cite{wu2019sgc}. At each incremental session, it creates a dedicated expert for the newly introduced classes and freezes the expert after training. In this way, subsequent learning does not directly overwrite the decision functions acquired in previous sessions.
To perform expert selection without task identities, MAEM equips each expert with an independent Masked Autoencoder (MAE) \cite{he2022mae,hou2022graphmae}. During inference, all experts reconstruct the representation of each test node, and the node is assigned to the expert with the lowest reconstruction error. The MAE reconstruction error therefore serves as an unbiased routing signal. Unlike a shared learnable router, this mechanism requires neither the session identity of the test node nor continual router fine-tuning, thereby mitigating recency bias at the routing level.
MAEM further controls model growth by maintaining a predefined expert capacity. Once this capacity is reached, selected experts are consolidated into a single expert. Git Re-Basin first aligns their hidden neurons and performs an initial weight-space fusion \cite{ainsworth2023gitrebasin}. A DeepInversion-based procedure then leverages the distributional knowledge retained in the MAEs to generate topology-aware pseudo data for further distilling the original experts into the fused expert \cite{yin2020deepinversion}. This process preserves graph-related knowledge without storing or revisiting historical nodes and keeps the number of experts bounded throughout continual learning.
Our main contributions are summarized as follows:
\begin{itemize}
    \item We propose MAEM, a replay-free dynamic MoE framework that mitigates recency bias through isolated and frozen session-specific experts.
    \item We introduce an MAE-based routing mechanism that uses reconstruction error as an unbiased routing signal, requiring neither task identities nor a continually fine-tuned shared router.
    \item We develop a capacity-controlled expert fusion strategy that combines Git Re-Basin with topology-aware pseudo data generated by DeepInversion-based synthesis, enabling knowledge consolidation without historical graph data.
    \item Experiments on four graph class-incremental benchmarks demonstrate that MAEM outperforms competitive baselines while maintaining bounded model capacity.
\end{itemize}

\section{Preliminaries}

\subsection{Class-Incremental Graph Learning}
Let $\mathcal{G}=(\mathcal{V},\mathcal{E},\mathbf{X})$ denote a graph, where $\mathcal{V}$ and $\mathcal{E}$ are the node and edge sets, and $\mathbf{X}$ contains the node features. Each node $v\in\mathcal{V}$ is associated with a class label $y_v$.
The learning process consists of a sequence of sessions $\{\mathcal{T}_1,\ldots,\mathcal{T}_K\}$ under the class-incremental graph learning setting \cite{zhang2022cglb}. Session $t$ introduces a set of previously unseen classes $\mathcal{C}_t$, where $\mathcal{C}_i\cap\mathcal{C}_j=\varnothing$ for $i\neq j$. The classes observed through session $t$ are denoted by $\mathcal{C}_{1:t}=\bigcup_{i=1}^{t}\mathcal{C}_i$. Correspondingly, $\mathcal{V}_t=\{v\in\mathcal{V}\mid y_v\in\mathcal{C}_t\}$ denotes the nodes introduced in session $t$.
At session $t$, the learner observes the cumulative graph
\[
\begin{aligned}
\mathcal{G}_{1:t}
    &= (\mathcal{V}_{1:t},\mathcal{E}_{1:t},\mathbf{X}_{1:t}), \\
\mathcal{V}_{1:t}
    &= \bigcup_{i=1}^{t}\mathcal{V}_i, \quad
\mathcal{E}_{1:t}
    = \{(u,v)\in\mathcal{E}\mid
        u,v\in\mathcal{V}_{1:t}\}.
\end{aligned}
\]
Therefore, the graph retains edges both within and across incremental sessions, rather than treating each session as an isolated subgraph \cite{zhang2022cglb}.
Under the replay-free setting, only the training nodes of the current session, denoted by $\mathcal{V}_t^{\mathrm{tr}}$, provide supervision. Historical labels are not reused in later sessions, although previously observed nodes may remain in the cumulative graph to provide structural context.
After session $t$, the model must be able to classify test nodes among all observed classes in $\mathcal{C}_{1:t}$. The session identity of each test node is unavailable \cite{vandeven2019three,zhang2022cglb}.

\subsection{Simplified Graph Convolution}
MAEM employs Simplified Graph Convolution (SGC) to obtain graph-aware node representations \cite{wu2019sgc}. Let $\mathbf{A}_t$ be the adjacency matrix of $\mathcal{G}_{1:t}$, $\widetilde{\mathbf{A}}_t=\mathbf{A}_t+\mathbf{I}$ be its self-loop augmented form, and $\widetilde{\mathbf{D}}_t$ be the corresponding degree matrix. The normalized propagation matrix is $\widehat{\mathbf{A}}_t=\widetilde{\mathbf{D}}_t^{-1/2}\widetilde{\mathbf{A}}_t\widetilde{\mathbf{D}}_t^{-1/2}$.
SGC removes intermediate nonlinear transformations and propagates node features for $L$ steps:
\[
\mathbf{H}_t=\widehat{\mathbf{A}}_t^{\,L}\mathbf{X}_{1:t}.
\]

The resulting representation $\mathbf{H}_t$ captures neighborhood information and serves as the common input to all experts. This separates graph propagation from session-specific expert optimization.

\section{Methodology}

\begin{figure*}[t]
    \centering
    \includegraphics[width=0.95\textwidth]{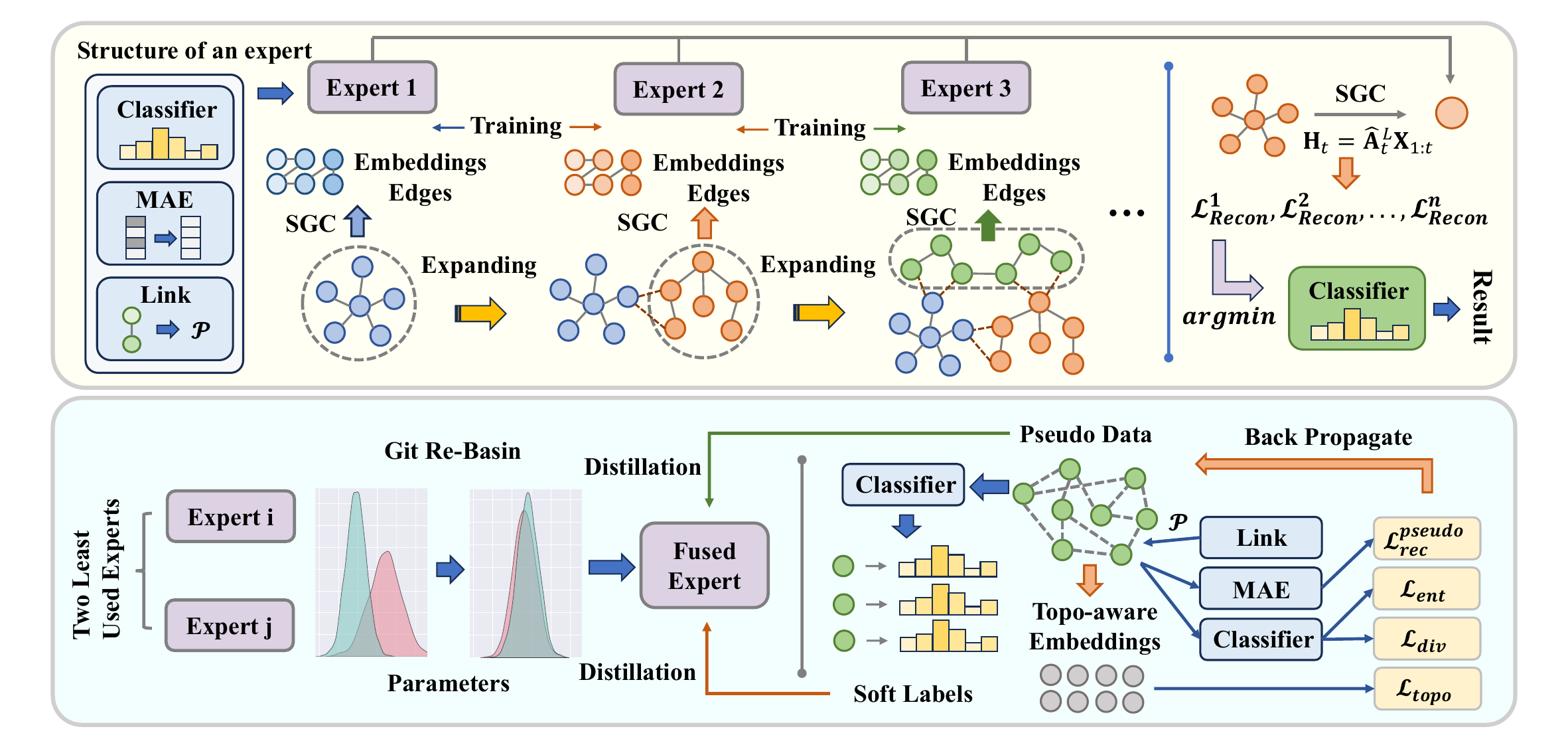}
    \caption{Overall framework of MAEM.}
    \label{fig:framework}
\end{figure*}

\subsection{Design Principles and Overall Framework}

MAEM is designed to mitigate recency bias in replay-free class-incremental graph learning while maintaining a controllable model size. It follows three principles.

The overall framework of MAEM is illustrated in Figure~\ref{fig:framework}.

First, MAEM employs SGC \cite{wu2019sgc} to separate graph propagation from expert optimization. Unlike a shared graph encoder that is repeatedly updated on new classes, SGC does not introduce session-dependent trainable transformations, thereby alleviating optimization-induced feature drift.

Second, MAEM creates and freezes a dedicated expert for each incremental session. Each expert contains a classifier, a masked autoencoder (MAE) \cite{he2022mae,hou2022graphmae}, and a neighbor predictor. At inference, MAE reconstruction errors are used to select an expert without requiring the session identity of the test node or continually fine-tuning a shared router. This prevents the routing mechanism from developing an explicit preference toward recently added experts.

Third, MAEM maintains a predefined expert capacity. Once this capacity is reached, selected experts are first aligned and merged through Git Re-Basin \cite{ainsworth2023gitrebasin}. Topology-aware pseudo representations are then generated through a DeepInversion-inspired procedure and used to further distill the original knowledge into the fused expert. This enables MAEM to control model growth without replaying historical graph data.

\subsection{Dynamic Expert Learning and MAE-Based Routing}

At session $t$, MAEM creates an expert
$\mathcal{E}_t=\{c_t,f_t,q_t\}$,
where $c_t$ is the classifier for the classes in $\mathcal{C}_t$, $f_t$ is the masked autoencoder used for routing, and $q_t$ is the neighbor predictor used during expert fusion. The expert is trained using the current-session nodes and is subsequently frozen.

\paragraph{Feature-level masked autoencoder.}
For a node representation $\mathbf{h}\in\mathbb{R}^d$ produced by SGC, we sample a binary mask $\mathbf{M}\in\{0,1\}^d$. Let $\mathbf{m}_t$ denote the learnable mask token of expert $t$. The corrupted representation is
$\widehat{\mathbf{h}}
=
\mathbf{M}\odot\mathbf{h}
+
(\mathbf{1}-\mathbf{M})\odot\mathbf{m}_t$.
The MAE reconstructs the complete representation through
\[
f_t(\widehat{\mathbf{h}})
=
\mathbf{W}_{t,2}
\sigma\!\left(
\mathbf{W}_{t,1}\widehat{\mathbf{h}}
+
\mathbf{b}_{t,1}
\right)
+
\mathbf{b}_{t,2},
\]
where $\sigma$ denotes ReLU. For theoretical clarity, we formulate MAE training using mean squared reconstruction error:
\[
\mathcal{L}_{\mathrm{rec}}^{(t)}
=
\frac{1}{|\mathcal{V}_t^{\mathrm{tr}}|}
\sum_{v\in\mathcal{V}_t^{\mathrm{tr}}}
\left\|
\mathbf{h}_v
-
f_t(\widehat{\mathbf{h}}_v)
\right\|_2^2.
\]
In practice, we use a direction-sensitive reconstruction loss that reduces the dominance of feature magnitude while retaining exact reconstruction as its pointwise optimum. The practical loss is used consistently for MAE training, inference routing, and pseudo-data generation, and its formulation is provided in Appendix~\ref{app:mae-loss}.

MAEM performs masking and reconstruction in the SGC representation space rather than reconstructing raw node attributes through a graph encoder as in GraphMAE \cite{hou2022graphmae}. This choice directly supports replay-free expert fusion. Synthesizing historical raw graphs would require jointly generating node attributes and discrete graph structures. In contrast, MAEM treats each SGC representation as a basic generation unit and optimizes pseudo representations directly in a continuous space. Their structural characteristics are subsequently preserved by topology-aware constraints supplied by the neighbor predictor.

\paragraph{Reconstruction-based expert routing.}
For a test representation $\mathbf{h}$, the same mask realization is applied to all experts. The reconstruction score of expert $e$ is
\[
R_e(\mathbf{h})
=
\left\|
\mathbf{h}
-
f_e(\widehat{\mathbf{h}})
\right\|_2^2.
\]
The selected expert is
\[
e^*(\mathbf{h})
=
\arg\min_{e\in\{1,\ldots,E\}}
R_e(\mathbf{h}),
\]
where $E$ is the current number of experts. The selected expert then classifies $\mathbf{h}$ using its expert-specific classifier.

This routing mechanism requires neither a known session identifier nor a shared router that is continually updated on new sessions. All experts are evaluated using the same reconstruction criterion. Therefore, the routing criterion is recency-unbiased in the sense that it contains no shared routing parameter optimized preferentially on recent-session data.

The independent learning and routing of frozen experts further provide an order-invariance property when cross-session representation changes and expert fusion are excluded.

\begin{theorem}[Order Invariance without Expert Fusion]
\label{thm:order_invariance}
Consider a fixed collection of class-incremental sessions
$\{\mathcal{T}_1,\ldots,\mathcal{T}_K\}$ and any two permutations
$\pi$ and $\rho$ of their arrival order. Assume that the influence of
cross-session edges is ignored, such that the SGC representation
$\mathbf{H}_k$ of session $\mathcal{T}_k$ depends only on its
session-specific graph and is independent of the preceding sessions.
If expert fusion is disabled and every session-specific objective is
optimized to its functionally unique task-wise solution, then the
final expert collections obtained under $\pi$ and $\rho$ are
identical up to a permutation of expert indices. Furthermore, for
every input with a unique minimum-reconstruction-error expert, the
routing decision and final global prediction are invariant to the
session order.
\end{theorem}

The proof is provided in Appendix~\ref{app:proof-order-invariance}. Theorem~\ref{thm:order_invariance} shows that frozen independent experts do not intrinsically depend on the order in which sessions are presented. In the complete setting, order dependence may still arise from representation changes induced by cross-session edges and from capacity-triggered expert fusion.

We next analyze the reliability of reconstruction-error-based routing.

\begin{theorem}[Routing Reliability]
\label{thm:routing}
Let $\mathcal{D}_i$ denote the representation distribution associated
with expert $i$, and let $\widehat{\mathbf{h}}$ be a masked observation
of $\mathbf{h}\sim\mathcal{D}_i$. If every MAE is optimized under
expected MSE, then
$f_i^*(\widehat{\mathbf{h}})
=
\mathbb{E}_{\mathcal{D}_i}
[\mathbf{h}\mid\widehat{\mathbf{h}}]$.
For every competing expert $j\neq i$, define
$\boldsymbol{\Delta}_{ij}
=
f_i^*(\widehat{\mathbf{h}})
-
f_j^*(\widehat{\mathbf{h}})$,
$\boldsymbol{\Sigma}_i
=
\operatorname{Cov}
[\mathbf{h}\mid\widehat{\mathbf{h}},\mathcal{D}_i]$,
$m_{ij}
=
\|\boldsymbol{\Delta}_{ij}\|_2^2$, and
$v_{ij}
=
4\boldsymbol{\Delta}_{ij}^{\top}
\boldsymbol{\Sigma}_i
\boldsymbol{\Delta}_{ij}$.
If $m_{ij}>0$, the pairwise routing probability satisfies
\[
\Pr(
R_i<R_j
\mid\widehat{\mathbf{h}},\mathcal{D}_i)
\ge
\frac{m_{ij}^2}{m_{ij}^2+v_{ij}}.
\]
Moreover, if $m_{ij}>0$ for every
$j\in\mathcal{S}\setminus\{i\}$, then
\[
\begin{aligned}
&\Pr\!\left(
R_i<
\min_{j\in\mathcal{S}\setminus\{i\}}R_j
\mid\widehat{\mathbf{h}},\mathcal{D}_i
\right) \\
&\qquad\ge
\left[
1-
\sum_{j\in\mathcal{S}\setminus\{i\}}
\frac{v_{ij}}{m_{ij}^2+v_{ij}}
\right]_+,
\end{aligned}
\]
where $[z]_+=\max(z,0)$.
\end{theorem}

The proof is provided in Appendix~\ref{app:proof-routing}. Theorem~\ref{thm:routing} shows that routing becomes more reliable when the conditional reconstructions of different experts are well separated and the irreducible reconstruction variance is small. In particular, when the masked dimensions can be accurately inferred from the observed dimensions and the expert-specific reconstruction gap is nonzero, the routing probability approaches one.

\subsection{Capacity-Controlled Expert Fusion}

When the number of experts reaches the capacity $B$, MAEM consolidates two least-activated experts, denoted by $\mathcal{E}_a$ and $\mathcal{E}_b$, according to the accumulated routing statistics. The fusion contains two stages: permutation-aligned parameter merging and topology-aware data-free distillation.

\paragraph{Neighbor predictor.}
Each expert contains a neighbor predictor $q_e$ to preserve structural information during pseudo-data generation. For two representations $\mathbf{h}_u$ and $\mathbf{h}_v$, we construct the symmetric pair representation
$\mathbf{z}_{uv}
=
\operatorname{Concat}
\left(
\mathbf{h}_u+\mathbf{h}_v,\,
\mathbf{h}_u\odot\mathbf{h}_v
\right)$.
The probability that the corresponding nodes are connected is
$p_{uv}^{(e)}
=
\operatorname{sigmoid}
\left(q_e(\mathbf{z}_{uv})\right)$.
The neighbor predictor is trained using binary cross-entropy, with real edges between current-session training nodes as positive pairs and randomly sampled non-edges as negative pairs.

After training expert $e$, the neighbor predictor identifies the $k$ most probable neighbors of each representation. The resulting topology-enriched representation is
\[
\widetilde{\mathbf{h}}_u^{(e)}
=
\mathbf{h}_u
+
\frac{1}{k}
\sum_{v\in\operatorname{TopK}_e(u)}
p_{uv}^{(e)}\mathbf{h}_v.
\]
MAEM retains only the mean and variance of these enriched representations:
$\boldsymbol{\mu}_e
=
\mathbb{E}
[\widetilde{\mathbf{h}}^{(e)}]$
and
$\boldsymbol{\nu}_e
=
\operatorname{Var}
[\widetilde{\mathbf{h}}^{(e)}]$.
These statistics provide topology-aware constraints for later pseudo-data generation without retaining historical nodes or edges.

\paragraph{Permutation-aligned parameter fusion.}
Directly averaging independently trained networks may combine hidden neurons with inconsistent permutations. MAEM therefore first aligns functionally corresponding neurons using Git Re-Basin \cite{ainsworth2023gitrebasin}.

For compatible parameters of experts $a$ and $b$, the alignment seeks a permutation
\[
P^*
=
\arg\min_{P\in\mathcal{P}}
\left\|
\boldsymbol{\theta}_a
-
P(\boldsymbol{\theta}_b)
\right\|_2^2,
\]
where $\mathcal{P}$ denotes the set of valid hidden-unit permutations. In practice, the permutation is obtained through layer-wise Hungarian matching.

Let $n_a$ and $n_b$ denote the numbers of training samples represented by the two experts, and define
$\lambda
=
\frac{n_a}{n_a+n_b}$.
The initial fused parameters are
$
\boldsymbol{\theta}_m
=
\lambda\boldsymbol{\theta}_a
+
(1-\lambda)P^*(\boldsymbol{\theta}_b).
$
For classifiers managing disjoint class sets, their hidden neurons are aligned while their output rows are mapped into the union of the two global class sets.

The following theorem explains why permutation alignment provides a favorable initialization for subsequent distillation.

\begin{theorem}[Output Discrepancy of Aligned Fusion]
\label{thm:fusion}
Let $P$ be a function-preserving neuron permutation and define
$
\boldsymbol{\theta}_m
=
\lambda\boldsymbol{\theta}_a
+
(1-\lambda)P(\boldsymbol{\theta}_b).
$
The corresponding weighted ensemble output is
$
f_{\mathrm{ens}}(\mathbf{x})
=
\lambda f_{\boldsymbol{\theta}_a}(\mathbf{x})
+
(1-\lambda)
f_{P(\boldsymbol{\theta}_b)}(\mathbf{x}).
$
If $\boldsymbol{\theta}\mapsto
f_{\boldsymbol{\theta}}(\mathbf{x})$ is locally
$L_{\mathbf{x}}$-Lipschitz on the fusion segment, then
\[
\begin{aligned}
&
\left\|
f_{\mathrm{ens}}(\mathbf{x})
-
f_{\boldsymbol{\theta}_m}(\mathbf{x})
\right\|_2 \\
&\qquad\le
2L_{\mathbf{x}}\lambda(1-\lambda)
\left\|
\boldsymbol{\theta}_a
-
P(\boldsymbol{\theta}_b)
\right\|_2.
\end{aligned}
\]
If its Jacobian is additionally locally
$\beta_{\mathbf{x}}$-Lipschitz, then
\[
\begin{aligned}
&
\left\|
f_{\mathrm{ens}}(\mathbf{x})
-
f_{\boldsymbol{\theta}_m}(\mathbf{x})
\right\|_2 \\
&\qquad\le
\frac{\beta_{\mathbf{x}}}{2}
\lambda(1-\lambda)
\left\|
\boldsymbol{\theta}_a
-
P(\boldsymbol{\theta}_b)
\right\|_2^2.
\end{aligned}
\]
\end{theorem}

The proof is provided in Appendix~\ref{app:proof-fusion}. Theorem~\ref{thm:fusion} does not imply that parameter averaging alone preserves all expert knowledge. Instead, it shows that neuron alignment reduces an upper bound on the discrepancy between the fused model and the corresponding weighted ensemble, providing a suitable initialization for subsequent distillation.

\paragraph{Topology-aware pseudo-data generation.}
Parameter fusion alone may still lose task-specific knowledge, particularly because the original classifiers manage different class sets. MAEM therefore generates pseudo representations separately from the two frozen experts.

For expert $e$, a pseudo batch
$\mathbf{Z}_e=\{\mathbf{z}_n\}_{n=1}^{N_e}$
is initialized from Gaussian noise and optimized while the expert parameters remain frozen. Let
$\mathbf{p}_n^{(e)}
=
\operatorname{softmax}
(c_e(\mathbf{z}_n))$
be the classifier prediction for pseudo representation $\mathbf{z}_n$. The generation objective consists of four terms.

The reconstruction loss encourages pseudo representations to lie in regions modeled by the expert:
$\mathcal{L}_{\mathrm{rec}}^{\mathrm{pseudo}}
=
\frac{1}{N_e}
\sum_{n=1}^{N_e}
\left\|
\mathbf{z}_n
-
f_e(\widehat{\mathbf{z}}_n)
\right\|_2^2$.

The prediction entropy is minimized to produce class-confident representations:
$\mathcal{L}_{\mathrm{ent}}
=
-\frac{1}{N_e}
\sum_{n=1}^{N_e}
\sum_c
p_{n,c}^{(e)}
\log p_{n,c}^{(e)}$.

To prevent the pseudo batch from collapsing into a small subset of classes, we define
$\overline{\mathbf{p}}^{(e)}
=
\frac{1}{N_e}
\sum_{n=1}^{N_e}
\mathbf{p}_n^{(e)}$
and minimize
$\mathcal{L}_{\mathrm{div}}
=
\sum_c
\overline{p}_c^{(e)}
\log\overline{p}_c^{(e)}$.

The frozen neighbor predictor then constructs topology-enriched pseudo representations. Let their mean and variance be
$\boldsymbol{\mu}_{\mathrm{pseudo}}^{(e)}$
and
$\boldsymbol{\nu}_{\mathrm{pseudo}}^{(e)}$.
The topology-aware loss is
\[
\begin{aligned}
\mathcal{L}_{\mathrm{topo}}
={}&
\left\|
\boldsymbol{\mu}_{\mathrm{pseudo}}^{(e)}
-
\boldsymbol{\mu}_e
\right\|_2^2 \\
&+
\left\|
\boldsymbol{\nu}_{\mathrm{pseudo}}^{(e)}
-
\boldsymbol{\nu}_e
\right\|_2^2.
\end{aligned}
\]

The complete pseudo-data generation objective is
$\mathcal{L}_{\mathrm{gen}}
=
\mathcal{L}_{\mathrm{rec}}^{\mathrm{pseudo}}
+
\omega_{\mathrm{ent}}
\mathcal{L}_{\mathrm{ent}}
+
\omega_{\mathrm{div}}
\mathcal{L}_{\mathrm{div}}
+
\omega_{\mathrm{topo}}
\mathcal{L}_{\mathrm{topo}}$.

This procedure follows the input-optimization principle of DeepInversion \cite{yin2020deepinversion} but generates continuous SGC representations rather than raw nodes or complete graphs. The topology-aware term preserves graph-related characteristics without explicitly synthesizing a discrete adjacency matrix.

\paragraph{Knowledge distillation.}
After pseudo-data generation, the pseudo batches of experts $a$ and $b$ are combined. The original classifiers produce temperature-softened targets \cite{hinton2015distilling} :
$\mathbf{s}_e(\mathbf{z})
=
\operatorname{softmax}
\left(
\frac{c_e(\mathbf{z})}{T}
\right)$.
Since the two experts may manage different classes, their outputs are mapped into the unified class space of the fused expert. The fused classifier is trained using
$\mathcal{L}_{\mathrm{KD}}^{\mathrm{cls}}
=
-T^2
\sum_{\mathbf{z}}
\mathbf{s}_e(\mathbf{z})^\top
\log
\operatorname{softmax}
\left(
\frac{c_m(\mathbf{z})}{T}
\right)$.

The fused MAE is further trained to reconstruct the combined pseudo representations. The fused neighbor predictor is distilled using the soft edge probabilities produced by the original neighbor predictors on pseudo-representation pairs.

After distillation, experts $\mathcal{E}_a$ and $\mathcal{E}_b$ are replaced by the fused expert $\mathcal{E}_m$. The expert pool is thereby reduced by one before the newly arriving expert is added, ensuring that the number of active experts never exceeds the predefined capacity $B$.

\section{Experiments}

\subsection{Experimental Settings}

\textbf{Datasets.}
We evaluate MAEM on four node-classification benchmarks:
\textsc{ogbn-arxiv} \cite{hu2020ogb},
\textsc{Cora-Full} \cite{bojchevski2018deep},
\textsc{Coauthor-CS} \cite{shchur2018pitfalls}, and
\textsc{WikiCS} \cite{mernyei2020wikics}.
These datasets cover citation, co-authorship, and hyperlink networks
with different scales and class distributions. All datasets follow the
class-incremental setting with inter-session edges preserved
\cite{zhang2022cglb}. Detailed statistics and data protocols are
provided in Appendix~\ref{app:datasets-metrics}.

\textbf{Baselines.}
We compare with sequential fine-tuning (\emph{Bare});
EWC, MAS, LwF, and TWP
\cite{kirkpatrick2017ewc,aljundi2018mas,li2017lwf,liu2021twp};
replay-based ER-GNN, CaT, DeLoMe, PDGNNs, and GSIP
\cite{zhou2021ergnn,liu2023cat,niu2024delome,zhang2024pdgnn,li2024gsip};
and the parameter-isolation method TAAM \cite{liu2026taam}.
Further descriptions are provided in Appendix~\ref{app:baselines}.

\textbf{Evaluation Metrics.}
We report final cumulative Micro accuracy, Average Performance (AP),
and Average Forgetting (AF) following CGLB \cite{zhang2022cglb}.
Higher Micro and AP are better, while AF closer to zero indicates less
forgetting. Formal definitions are provided in
Appendix~\ref{app:datasets-metrics}.

\subsection{Main Results}

\begin{table*}[t]
\centering
{\small
\setlength{\tabcolsep}{2pt}
\begin{tabular}{lccccccc}
\toprule
& & \multicolumn{3}{c}{OGBN-Arxiv}
& \multicolumn{3}{c}{Cora-Full} \\
\cmidrule(lr){3-5}\cmidrule(lr){6-8}
Method & Replay
& Micro & AP & AF
& Micro & AP & AF \\
\midrule
Bare & $\times$
& $0.52\pm0.24$ & $6.99\pm0.46$ & $-83.71\pm0.87$
& $1.60\pm0.16$ & $6.86\pm0.48$ & $-90.99\pm0.69$ \\
EWC & $\times$
& $0.39\pm0.01$ & $6.90\pm0.17$ & $-75.86\pm1.59$
& $2.02\pm0.31$ & $7.38\pm0.62$ & $-90.59\pm0.64$ \\
MAS & $\times$
& $0.37\pm0.01$ & $6.61\pm0.20$ & $-75.58\pm1.16$
& $1.18\pm0.03$ & $5.53\pm0.13$ & $-91.26\pm0.30$ \\
LwF & $\times$
& $0.72\pm0.22$ & $6.78\pm0.35$ & $-81.97\pm1.84$
& $2.03\pm0.17$ & $7.99\pm0.55$ & $-87.92\pm0.96$ \\
TWP & $\times$
& $0.37\pm0.02$ & $6.54\pm0.30$ & $-76.68\pm1.41$
& $1.10\pm0.03$ & $4.89\pm0.11$ & $-87.44\pm0.56$ \\
\midrule
ER-GNN & $\checkmark$
& $0.42\pm0.01$ & $7.32\pm0.19$ & $-83.16\pm0.95$
& $1.82\pm0.30$ & $6.70\pm0.47$ & $-89.08\pm0.30$ \\
CaT & $\checkmark$
& $30.09\pm0.35$ & $39.51\pm0.31$ & $\underline{-17.82\pm0.38}$
& $55.56\pm1.37$ & $52.36\pm0.27$ & $-11.45\pm1.05$ \\
DeLoMe & $\checkmark$
& $11.64\pm0.25$ & $34.39\pm0.68$ & $-53.70\pm0.85$
& $57.96\pm0.11$ & $61.41\pm0.19$ & $-24.02\pm0.36$ \\
PDGNNs & $\checkmark$
& $43.48\pm0.89$ & $\mathbf{48.94\pm0.48}$ & $-31.88\pm0.97$
& $51.72\pm0.54$ & $54.59\pm0.51$ & $-36.35\pm0.27$ \\
GSIP & $\checkmark$
& $3.45\pm0.97$ & $10.03\pm0.71$ & $-76.14\pm1.44$
& $23.59\pm1.00$ & $24.35\pm0.65$ & $-71.76\pm0.70$ \\
TAAM & $\times$
& $31.29\pm0.01$ & $29.54\pm0.04$ & $-25.45\pm0.03$
& $41.08\pm0.12$ & $38.04\pm0.11$ & $-15.91\pm0.19$ \\
\midrule
MAEM$^{\mathsf{N}}$ & $\times$
& $\mathbf{50.34\pm1.41}$ & $\underline{46.35\pm0.27}$ & $\mathbf{-15.04\pm0.68}$
& $\mathbf{68.22\pm0.08}$ & $\mathbf{65.54\pm0.25}$ & $\mathbf{-7.85\pm0.36}$ \\
MAEM$^{\mathsf{M}}$ & $\times$
& $\underline{48.51\pm1.85}$ & $41.97\pm0.86$ & $-19.53\pm2.60$
& $\underline{67.40\pm0.28}$ & $\underline{63.04\pm0.25}$ & $\underline{-10.84\pm0.36}$ \\
\midrule[\heavyrulewidth]
& & \multicolumn{3}{c}{Coauthor-CS}
& \multicolumn{3}{c}{WikiCS} \\
\cmidrule(lr){3-5}\cmidrule(lr){6-8}
Method & Replay
& Micro & AP & AF
& Micro & AP & AF \\
\midrule
Bare & $\times$
& $4.71\pm0.32$ & $14.80\pm0.53$ & $-96.96\pm0.57$
& $6.53\pm0.21$ & $20.20\pm0.20$ & $-90.38\pm0.91$ \\
EWC & $\times$
& $5.91\pm0.75$ & $17.07\pm1.45$ & $-94.39\pm1.76$
& $6.02\pm0.02$ & $19.74\pm0.06$ & $-87.69\pm3.91$ \\
MAS & $\times$
& $4.33\pm0.00$ & $14.20\pm0.00$ & $-96.83\pm0.12$
& $6.00\pm0.08$ & $19.65\pm0.27$ & $-88.74\pm2.34$ \\
LwF & $\times$
& $6.14\pm0.99$ & $19.38\pm2.86$ & $-91.82\pm3.33$
& $6.93\pm1.72$ & $19.73\pm0.90$ & $-88.22\pm1.03$ \\
TWP & $\times$
& $4.20\pm0.07$ & $13.76\pm0.22$ & $-96.50\pm0.28$
& $6.03\pm0.02$ & $19.77\pm0.07$ & $-89.12\pm1.24$ \\
\midrule
ER-GNN & $\checkmark$
& $44.14\pm9.92$ & $47.59\pm11.51$ & $-56.76\pm13.35$
& $11.41\pm0.69$ & $23.57\pm0.54$ & $-85.65\pm0.83$ \\
CaT & $\checkmark$
& $90.24\pm1.56$ & $88.37\pm0.93$ & $\underline{-4.50\pm1.32}$
& $\underline{74.86\pm0.76}$ & $\underline{74.36\pm0.59}$ & $\mathbf{-6.54\pm0.83}$ \\
DeLoMe & $\checkmark$
& $82.52\pm0.78$ & $84.04\pm0.77$ & $-16.09\pm0.86$
& $59.83\pm0.80$ & $68.49\pm0.52$ & $-32.20\pm0.78$ \\
PDGNNs & $\checkmark$
& $88.02\pm0.93$ & $\underline{89.05\pm0.47}$ & $-10.64\pm0.63$
& $63.47\pm3.20$ & $70.87\pm2.28$ & $-29.50\pm3.01$ \\
GSIP & $\checkmark$
& $36.26\pm4.29$ & $43.69\pm4.94$ & $-62.30\pm5.93$
& $17.32\pm2.43$ & $30.75\pm1.91$ & $-76.03\pm1.62$ \\
TAAM & $\times$
& $86.86\pm0.04$ & $85.04\pm0.07$ & $-7.32\pm0.04$
& $60.04\pm0.09$ & $64.09\pm0.10$ & $-21.17\pm0.24$ \\
\midrule
MAEM$^{\mathsf{N}}$ & $\times$
& $\mathbf{91.99\pm0.16}$ & $\mathbf{90.83\pm0.25}$ & $\mathbf{-4.24\pm0.21}$
& $\mathbf{75.42\pm1.64}$ & $\mathbf{76.84\pm0.73}$ & $\underline{-10.14\pm0.94}$ \\
MAEM$^{\mathsf{M}}$ & $\times$
& $\underline{91.87\pm0.19}$ & $89.71\pm0.41$ & $-5.49\pm0.46$
& $69.19\pm2.43$ & $69.29\pm1.94$ & $-20.08\pm2.28$ \\
\bottomrule
\end{tabular}
}
\caption{Main results on four class-incremental graph benchmarks.
Results are percentages reported as mean $\pm$ standard deviation over
five runs. $\checkmark$ and $\times$ indicate whether historical
information is replayed. MAEM$^{\mathsf{N}}$ and
MAEM$^{\mathsf{M}}$ denote MAEM without and with expert merging,
respectively. Bold and underlined values indicate the best and
second-best results.}
\label{tab:main_results}
\end{table*}

\paragraph{Overall performance.}
Table~\ref{tab:main_results} shows that MAEM$^{\mathsf{N}}$ achieves
the best Micro accuracy on all four datasets. It outperforms the
strongest baselines by 6.86, 10.26, 1.75, and 0.56 percentage points
on OGBN-Arxiv, Cora-Full, Coauthor-CS, and WikiCS, respectively.
MAEM also obtains the best AP on the latter three datasets, with gains
of 4.13, 1.78, and 2.48 points, while ranking second on OGBN-Arxiv.
These results are achieved without retaining or replaying historical
graph data, showing that frozen experts effectively protect
session-specific decision functions from subsequent optimization.

\paragraph{Forgetting resistance.}
MAEM$^{\mathsf{N}}$ achieves the best AF on OGBN-Arxiv, Cora-Full,
and Coauthor-CS. Although CaT obtains a better AF on WikiCS, MAEM
provides higher Micro and AP, indicating a better balance between
knowledge retention and cumulative predictive performance. This also
shows that low forgetting alone does not necessarily imply strong
classification across all observed classes.

\paragraph{Effect of expert merging.}
MAEM$^{\mathsf{M}}$ bounds model capacity while retaining the
second-best Micro accuracy on OGBN-Arxiv, Cora-Full, and Coauthor-CS.
Compared with MAEM$^{\mathsf{N}}$, its Micro decreases by only 1.83,
0.82, and 0.12 points on these datasets. The larger decrease on WikiCS
suggests that frequent consolidation is more challenging when
expert-specific distributions overlap substantially.

\subsection{Impact of Expert Capacity}
\label{sec:expert_capacity}

We vary the maximum number of experts $M$ from 6 to 10 on OGBN-Arxiv
and from 9 to 17 on Cora-Full while keeping all other settings fixed.

\begin{figure}[t]
    \centering
    \includegraphics[width=\columnwidth]{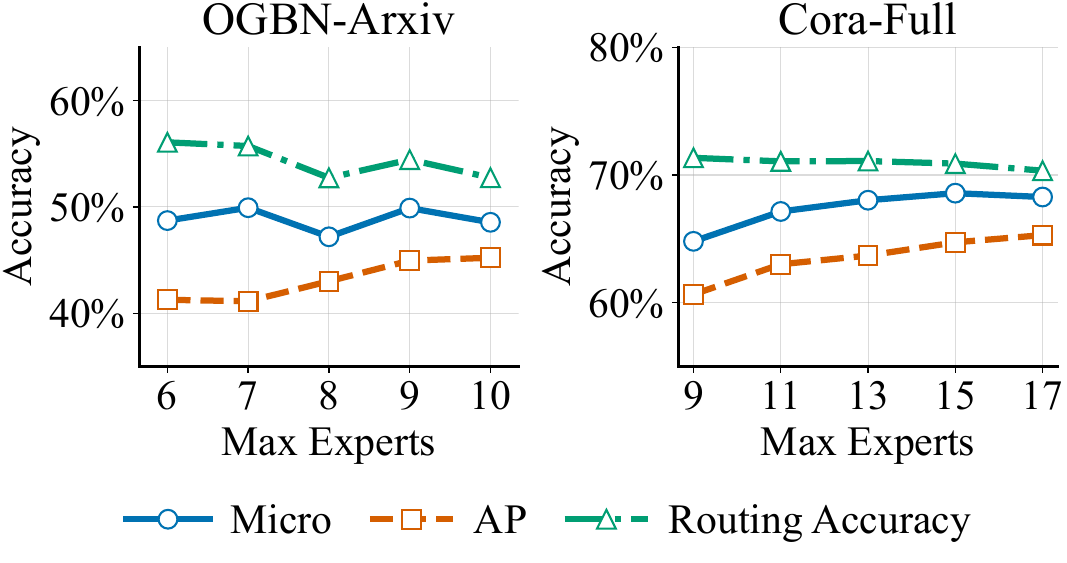}
    \caption{Final Micro accuracy, AP, and routing accuracy under
    different maximum numbers of experts on OGBN-Arxiv and Cora-Full.}
    \label{fig:expert_capacity}
\end{figure}

\begin{figure}[t]
    \centering
    \includegraphics[width=\columnwidth]{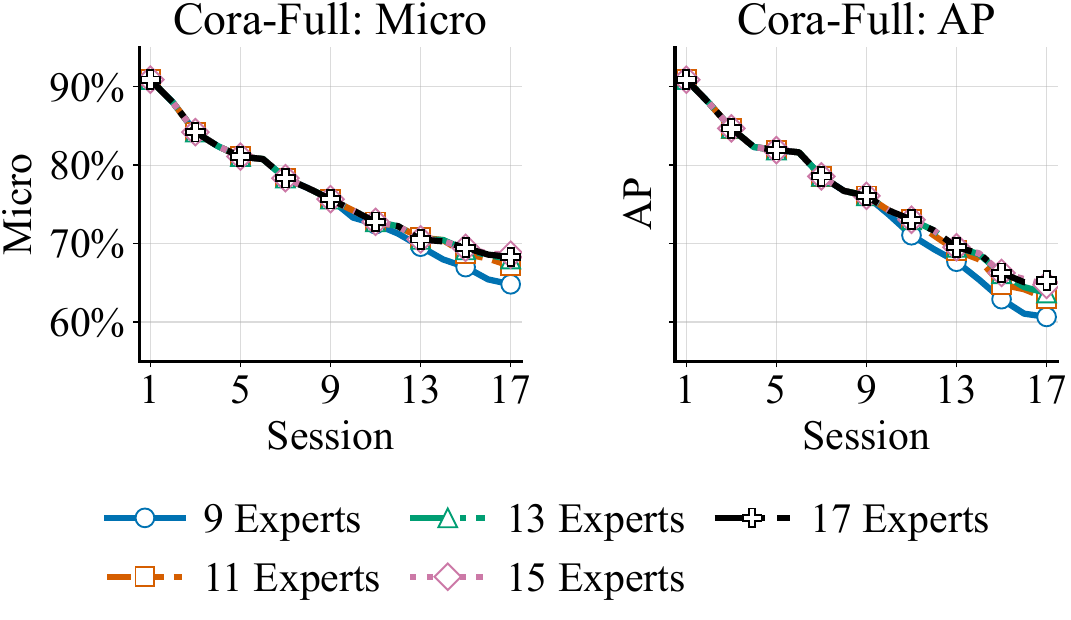}
    \caption{Session-wise Micro accuracy and AP on Cora-Full under
    different expert-capacity limits.}
    \label{fig:cora_dynamics}
\end{figure}

\paragraph{Final performance.}
Figure~\ref{fig:expert_capacity} shows that larger capacities generally
improve Micro and AP because fewer experts require consolidation.
On Cora-Full, AP increases from 60.65\% with 9 experts to 65.29\%
with 17 experts, while routing accuracy remains stable at approximately
70--71\%. On OGBN-Arxiv, AP also generally improves as more experts
are retained, despite moderate fluctuations in Micro and routing
accuracy. Thus, capacity primarily controls the frequency of knowledge
consolidation rather than the reliability of expert routing.

\paragraph{Learning dynamics.}
As shown in Figure~\ref{fig:cora_dynamics}, different capacity settings
behave similarly before fusion is triggered. Smaller capacities
subsequently lead to more frequent fusion and moderately lower
performance. Nevertheless, the curves decrease smoothly rather than
collapsing, demonstrating that MAEM can control model growth while
preserving substantial historical knowledge.

\subsection{Ablation Study}

\begin{table}[t]
\centering
\small
\setlength{\tabcolsep}{5pt}
\begin{tabular}{lcc}
\toprule
Variant & Micro & AP \\
\midrule
Full model
& $\mathbf{67.40\pm0.28}$
& $\mathbf{63.04\pm0.25}$ \\
Random initialization
& $61.13\pm0.30$
& $45.87\pm0.21$ \\
w/o distillation
& $60.21\pm0.39$
& $45.20\pm0.19$ \\
w/o topology-aware loss
& $60.77\pm0.30$
& $45.71\pm0.21$ \\
\bottomrule
\end{tabular}
\caption{Ablation results (\%) on Cora-Full.}
\label{tab:ablation}
\end{table}

Table~\ref{tab:ablation} shows that replacing aligned initialization
with random initialization and removing either distillation or the
topology-aware loss consistently reduce Micro and AP. In particular,
removing distillation causes the largest Micro reduction, while all
three variants exhibit substantial AP degradation. These results
confirm that aligned initialization, pseudo-data distillation, and
topology-aware synthesis all contribute to effective expert
consolidation.

\subsection{Further Experiments}

Further analyses of hyperparameters, backbone architectures, and
pseudo-data generation quality are provided in
Appendix~\ref{app:further-experiments}.

\section{Related Work}

\paragraph{Continual learning methods.}
Continual learning methods mainly use regularization, distillation, replay, or parameter isolation \cite{parisi2019continual,delange2022continual}. Representative approaches include EWC and MAS \cite{kirkpatrick2017ewc,aljundi2018mas}, LwF \cite{li2017lwf}, episodic replay \cite{rebuffi2017icarl,lopezpaz2017gem}, and task-specific modules \cite{rusu2016pnn,yoon2018den}. In continual graph learning, these strategies preserve topology-related parameters or replay nodes, condensed graphs, prototypes, and topology-aware embeddings \cite{zhang2022cglb,liu2021twp,zhou2021ergnn,liu2023cat,niu2024delome,zhang2024pdgnn,li2024gsip}. However, shared backbones remain susceptible to interference, replay retains historical information, and isolation may require task identities or unbounded growth. MAEM instead combines fixed graph propagation, independently frozen experts, and replay-free consolidation to reduce interference while controlling capacity.

\paragraph{Mixture-of-experts models.}
MoE models route inputs to specialized experts, enabling different components to capture distinct data patterns \cite{shazeer2017moe}. In continual learning, expert isolation reduces interference, while reconstruction errors can infer task identity without explicit labels \cite{aljundi2017expertgate}. Graph MoE methods capture structural heterogeneity using receptive-field-specific experts or sparse routing over evolving graph blocks \cite{wang2023gmoe,kong2025dymoe}. Nevertheless, learned routers may drift during sequential optimization, and continually expanding expert pools increase memory and inference costs. MAEM instead uses independent MAE reconstruction errors for task-identity-free routing and bounds long-term growth through expert fusion.

\paragraph{Data-free distillation and model fusion.}
Data-free distillation synthesizes surrogate inputs from frozen teachers when original data are unavailable, supporting knowledge transfer under storage or privacy constraints \cite{nayak2019zeroshot,chen2019datafree,yin2020deepinversion}. Model fusion combines independently trained networks in weight space, but permutation symmetries make naive averaging unreliable \cite{wortsman2022modelsoups,ainsworth2023gitrebasin}. MAEM combines Git Re-Basin alignment with topology-aware pseudo-feature distillation to merge experts with disjoint class spaces without retaining or revisiting historical graphs.

\section{Conclusion}
We proposed MAEM, a replay-free, capacity-controlled MoE framework for class-incremental graph learning. Frozen session experts and MAE-based routing mitigate recency bias without requiring task identities, while topology-aware data-free fusion bounds model growth without retaining historical graphs. Experiments on four benchmarks demonstrate strong performance and show that MAEM maintains bounded capacity with limited degradation.

\bibliography{aaai2027}

\clearpage
\appendix
\section{Proofs of Theoretical Results}
\label{app:proofs}

\subsection{Proof of Theorem~\ref{thm:order_invariance}}
\label{app:proof-order-invariance}

\paragraph{Proof.}
For every session $\mathcal{T}_k$, let
$\mathcal{J}_k(\boldsymbol{\phi}_k;\mathbf{H}_k)$ denote its complete
session-specific objective, where $\boldsymbol{\phi}_k$ contains the
parameters of its classifier, masked autoencoder, and neighbor
predictor.

By assumption, cross-session edges are ignored. Therefore, the SGC
representation $\mathbf{H}_k$ depends only on the session-specific
graph associated with $\mathcal{T}_k$ and is independent of which
sessions were presented before $\mathcal{T}_k$. Consequently, for
any two arrival orders $\pi$ and $\rho$, session $\mathcal{T}_k$
induces the same optimization problem:
\[
\boldsymbol{\phi}_k^*
\in
\arg\min_{\boldsymbol{\phi}_k}
\mathcal{J}_k(\boldsymbol{\phi}_k;\mathbf{H}_k).
\]

Expert fusion is disabled, and every expert is frozen after its
session is learned. Thus, optimizing the expert associated with
$\mathcal{T}_k$ neither reads nor modifies the parameters of any
other expert. The optimization problems of different sessions are
therefore independent.

Under arrival order $\pi$, the final ordered expert collection is
\[
\mathfrak{E}_{\pi}
=
\left(
\mathcal{E}_{\pi(1)}^*,
\ldots,
\mathcal{E}_{\pi(K)}^*
\right).
\]
Under arrival order $\rho$, it is
\[
\mathfrak{E}_{\rho}
=
\left(
\mathcal{E}_{\rho(1)}^*,
\ldots,
\mathcal{E}_{\rho(K)}^*
\right).
\]
Because every session-specific objective is assumed to have a
functionally unique task-wise solution, the expert learned for
$\mathcal{T}_k$ implements the same classifier, reconstruction
function, and neighbor predictor under both orders. Therefore,
$\mathfrak{E}_{\pi}$ and $\mathfrak{E}_{\rho}$ contain the same
experts and differ only in their numerical expert indices.

For a fixed input representation $\mathbf{h}$, let
\[
R_k(\mathbf{h})
=
\left\|
\mathbf{h}
-
f_k(\widehat{\mathbf{h}})
\right\|_2^2
\]
be the reconstruction score of the expert associated with
$\mathcal{T}_k$. Since the set of reconstruction functions is
identical under $\pi$ and $\rho$, the set of reconstruction scores
is also identical up to the same index permutation.

If the minimum reconstruction score is unique, both arrival orders
select the expert associated with the same session, although that
expert may occupy a different index in the expert pool. The selected
expert implements the same classifier and local-to-global class
mapping under both orders. Hence, its final global prediction is
also unchanged.

Therefore, the final expert collection is invariant up to an
expert-index permutation, and both the routing decision and global
prediction are invariant to session order.
\hfill$\square$

\subsection{Proof of Theorem~\ref{thm:routing}}
\label{app:proof-routing}

\paragraph{Proof.}
Condition on a fixed masked observation
$\widehat{\mathbf{h}}$ and the event
$\mathbf{h}\sim\mathcal{D}_i$. Define
\[
\boldsymbol{\mu}_i
=
\mathbb{E}_{\mathcal{D}_i}
[
\mathbf{h}\mid\widehat{\mathbf{h}}
]
\]
and
\[
\boldsymbol{\Sigma}_i
=
\operatorname{Cov}
[
\mathbf{h}\mid\widehat{\mathbf{h}},\mathcal{D}_i
].
\]

For any reconstruction function $g(\widehat{\mathbf{h}})$, the
conditional MSE can be decomposed as
\[
\begin{aligned}
&
\mathbb{E}
\left[
\left\|
\mathbf{h}
-
g(\widehat{\mathbf{h}})
\right\|_2^2
\mid
\widehat{\mathbf{h}},\mathcal{D}_i
\right]
\\
&=
\operatorname{tr}(\boldsymbol{\Sigma}_i)
+
\left\|
\boldsymbol{\mu}_i
-
g(\widehat{\mathbf{h}})
\right\|_2^2.
\end{aligned}
\]
The first term is independent of $g$, and the second is minimized
when
\[
g(\widehat{\mathbf{h}})
=
\boldsymbol{\mu}_i.
\]
Therefore, an expected-MSE-optimal MAE satisfies
\[
f_i^*(\widehat{\mathbf{h}})
=
\mathbb{E}_{\mathcal{D}_i}
[
\mathbf{h}\mid\widehat{\mathbf{h}}
].
\]

Define the residual
\[
\boldsymbol{\epsilon}
=
\mathbf{h}
-
f_i^*(\widehat{\mathbf{h}}).
\]
Conditioned on $\widehat{\mathbf{h}}$ and $\mathcal{D}_i$,
\[
\mathbb{E}[\boldsymbol{\epsilon}]=\mathbf{0},
\qquad
\operatorname{Cov}(\boldsymbol{\epsilon})
=
\boldsymbol{\Sigma}_i.
\]

For a competing expert $j$, recall that
\[
\boldsymbol{\Delta}_{ij}
=
f_i^*(\widehat{\mathbf{h}})
-
f_j^*(\widehat{\mathbf{h}}).
\]
The reconstruction error of the correct expert is
\[
R_i
=
\|\boldsymbol{\epsilon}\|_2^2,
\]
whereas
\[
\begin{aligned}
R_j
&=
\left\|
\mathbf{h}
-
f_j^*(\widehat{\mathbf{h}})
\right\|_2^2 \\
&=
\left\|
\boldsymbol{\epsilon}
+
\boldsymbol{\Delta}_{ij}
\right\|_2^2 \\
&=
\|\boldsymbol{\epsilon}\|_2^2
+
2\boldsymbol{\epsilon}^{\top}\boldsymbol{\Delta}_{ij}
+
\|\boldsymbol{\Delta}_{ij}\|_2^2.
\end{aligned}
\]
Hence,
\[
R_j-R_i
=
2\boldsymbol{\epsilon}^{\top}\boldsymbol{\Delta}_{ij}
+
m_{ij}.
\]

Let
\[
Z_{ij}
=
2\boldsymbol{\epsilon}^{\top}\boldsymbol{\Delta}_{ij}.
\]
Its conditional mean is
\[
\mathbb{E}[Z_{ij}]
=
2\mathbb{E}[\boldsymbol{\epsilon}]^{\top}
\boldsymbol{\Delta}_{ij}
=
0,
\]
and its conditional variance is
\[
\begin{aligned}
\operatorname{Var}(Z_{ij})
&=
4\operatorname{Var}
\left(
\boldsymbol{\epsilon}^{\top}\boldsymbol{\Delta}_{ij}
\right) \\
&=
4\boldsymbol{\Delta}_{ij}^{\top}
\boldsymbol{\Sigma}_i
\boldsymbol{\Delta}_{ij} \\
&=
v_{ij}.
\end{aligned}
\]

The correct expert has a lower reconstruction error precisely when
\[
R_i<R_j
\quad\Longleftrightarrow\quad
Z_{ij}>-m_{ij}.
\]
By the one-sided Chebyshev--Cantelli inequality,
\[
\Pr(Z_{ij}\leq -m_{ij})
\leq
\frac{v_{ij}}{v_{ij}+m_{ij}^2}.
\]
Taking complements gives
\[
\Pr(
R_i<R_j
\mid
\widehat{\mathbf{h}},\mathcal{D}_i
)
\geq
\frac{m_{ij}^2}{m_{ij}^2+v_{ij}}.
\]

For multiple experts, define the pairwise routing-failure event
\[
\mathcal{F}_{ij}
=
\{R_j\leq R_i\}.
\]
Correct routing requires that none of these events occurs:
\[
\left\{
R_i<
\min_{j\in\mathcal{S}\setminus\{i\}}R_j
\right\}
=
\bigcap_{j\in\mathcal{S}\setminus\{i\}}
\mathcal{F}_{ij}^{c}.
\]
Using the union bound,
\[
\begin{aligned}
&
\Pr
\left(
\bigcup_{j\in\mathcal{S}\setminus\{i\}}
\mathcal{F}_{ij}
\mid
\widehat{\mathbf{h}},\mathcal{D}_i
\right)
\\
&\leq
\sum_{j\in\mathcal{S}\setminus\{i\}}
\frac{v_{ij}}{m_{ij}^2+v_{ij}}.
\end{aligned}
\]
Therefore,
\[
\begin{aligned}
&\Pr\left(
R_i<
\min_{j\in\mathcal{S}\setminus\{i\}}R_j
\mid
\widehat{\mathbf{h}},\mathcal{D}_i
\right)
\\
&\geq
\left[
1-
\sum_{j\in\mathcal{S}\setminus\{i\}}
\frac{v_{ij}}{m_{ij}^2+v_{ij}}
\right]_+.
\end{aligned}
\]
This proves the pairwise and multi-expert bounds.
\hfill$\square$

\subsection{Proof of Theorem~\ref{thm:fusion}}
\label{app:proof-fusion}

\paragraph{Proof.}
Let
\[
\boldsymbol{\theta}_b'
=
P(\boldsymbol{\theta}_b)
\]
and define
\[
\mathbf{d}
=
\boldsymbol{\theta}_a-\boldsymbol{\theta}_b'.
\]
The fused parameters satisfy
\[
\boldsymbol{\theta}_m
=
\lambda\boldsymbol{\theta}_a
+
(1-\lambda)\boldsymbol{\theta}_b'.
\]
Therefore,
\[
\boldsymbol{\theta}_a-\boldsymbol{\theta}_m
=
(1-\lambda)\mathbf{d}
\]
and
\[
\boldsymbol{\theta}_b'-\boldsymbol{\theta}_m
=
-\lambda\mathbf{d}.
\]

Assume first that the parameter-to-output mapping is locally
$L_{\mathbf{x}}$-Lipschitz on the complete fusion segment. By the
triangle inequality,
\[
\begin{aligned}
&
\left\|
f_{\mathrm{ens}}(\mathbf{x})
-
f_{\boldsymbol{\theta}_m}(\mathbf{x})
\right\|_2
\\
&\leq
\lambda
\left\|
f_{\boldsymbol{\theta}_a}(\mathbf{x})
-
f_{\boldsymbol{\theta}_m}(\mathbf{x})
\right\|_2
\\
&\quad+
(1-\lambda)
\left\|
f_{\boldsymbol{\theta}_b'}(\mathbf{x})
-
f_{\boldsymbol{\theta}_m}(\mathbf{x})
\right\|_2.
\end{aligned}
\]
Applying the Lipschitz condition gives
\[
\begin{aligned}
&
\left\|
f_{\mathrm{ens}}(\mathbf{x})
-
f_{\boldsymbol{\theta}_m}(\mathbf{x})
\right\|_2
\\
&\leq
\lambda L_{\mathbf{x}}
\left\|
\boldsymbol{\theta}_a-\boldsymbol{\theta}_m
\right\|_2
+
(1-\lambda)L_{\mathbf{x}}
\left\|
\boldsymbol{\theta}_b'-\boldsymbol{\theta}_m
\right\|_2
\\
&=
\lambda L_{\mathbf{x}}(1-\lambda)\|\mathbf{d}\|_2
+
(1-\lambda)L_{\mathbf{x}}\lambda\|\mathbf{d}\|_2
\\
&=
2L_{\mathbf{x}}\lambda(1-\lambda)\|\mathbf{d}\|_2.
\end{aligned}
\]
Substituting
$\mathbf{d}=\boldsymbol{\theta}_a-P(\boldsymbol{\theta}_b)$
proves the first inequality.

Now assume that the Jacobian with respect to parameters is locally
$\beta_{\mathbf{x}}$-Lipschitz in operator norm. A first-order Taylor
expansion around $\boldsymbol{\theta}_m$ gives
\[
f_{\boldsymbol{\theta}}(\mathbf{x})
=
f_{\boldsymbol{\theta}_m}(\mathbf{x})
+
J_m(\boldsymbol{\theta}-\boldsymbol{\theta}_m)
+
\mathbf{r}(\boldsymbol{\theta}),
\]
where
\[
\|\mathbf{r}(\boldsymbol{\theta})\|_2
\leq
\frac{\beta_{\mathbf{x}}}{2}
\|\boldsymbol{\theta}-\boldsymbol{\theta}_m\|_2^2.
\]

In the weighted ensemble, the first-order terms are proportional to
\[
\lambda
(\boldsymbol{\theta}_a-\boldsymbol{\theta}_m)
+
(1-\lambda)
(\boldsymbol{\theta}_b'-\boldsymbol{\theta}_m).
\]
By the definition of $\boldsymbol{\theta}_m$, this expression equals
zero. Hence, the linear Taylor terms cancel exactly, and only the
second-order remainder terms remain. It follows that
\[
\begin{aligned}
&
\left\|
f_{\mathrm{ens}}(\mathbf{x})
-
f_{\boldsymbol{\theta}_m}(\mathbf{x})
\right\|_2
\\
&\leq
\lambda
\frac{\beta_{\mathbf{x}}}{2}
(1-\lambda)^2\|\mathbf{d}\|_2^2
\\
&\quad+
(1-\lambda)
\frac{\beta_{\mathbf{x}}}{2}
\lambda^2\|\mathbf{d}\|_2^2
\\
&=
\frac{\beta_{\mathbf{x}}}{2}
\lambda(1-\lambda)
\|\mathbf{d}\|_2^2.
\end{aligned}
\]
Substituting the definition of $\mathbf{d}$ proves the second
inequality. Since $P$ is function-preserving, it changes only the
parameterization of expert $b$, not its represented function.
\hfill$\square$

\section{Datasets and Evaluation Metrics}
\label{app:datasets-metrics}

\subsection{Dataset Statistics}
\label{app:dataset-statistics}

We evaluate on four node-classification benchmarks spanning citation,
co-authorship, and hyperlink graphs:
\textsc{ogbn-arxiv}~\cite{hu2020ogb},
\textsc{Cora-Full}~\cite{bojchevski2018deep},
\textsc{Coauthor-CS}~\cite{shchur2018pitfalls}, and
\textsc{WikiCS}~\cite{mernyei2020wikics}.
Table~\ref{tab:app-dataset-stats} summarizes the statistics used in our
experiments. High-dimensional features on \textsc{Cora-Full} and
\textsc{Coauthor-CS} are reduced by Truncated SVD before training; the
remaining datasets keep their original feature dimensions.

\begin{table*}[t]
\centering
\small
\setlength{\tabcolsep}{3.5pt}
\begin{tabular}{lrrrrrr}
\toprule
Dataset & Nodes & Edges & Feature Dimension & Classes & Class per Session & Sessions \\
\midrule
\textsc{ogbn-arxiv}
& 169{,}343 & 1{,}166{,}243 & 128 & 40 & 4 & 10 \\
\textsc{Cora-Full}
& 19{,}793 & 65{,}311 & 512$^{\dagger}$ & 70 & 4 & 17 \\
\textsc{Coauthor-CS}
& 18{,}333 & 81{,}894 & 256$^{\dagger}$ & 15 & 2 & 7 \\
\textsc{WikiCS}
& 11{,}701 & 216{,}123 & 300 & 10 & 2 & 5 \\
\bottomrule
\end{tabular}
\caption{Dataset statistics. ``Feat.'' is the feature dimension after
preprocessing; $^{\dagger}$ denotes Truncated-SVD reduction
(original dimensions: 8{,}710 for \textsc{Cora-Full} and 6{,}805 for
\textsc{Coauthor-CS}). ``$C$/Sess.'' is the number of classes per
session. For \textsc{Coauthor-CS}, the final incomplete session is
discarded, so 14 of 15 classes are used.}
\label{tab:app-dataset-stats}
\end{table*}

\subsection{Class-Incremental Data Protocol}
\label{app:data-protocol}

All methods share the same class-incremental expanding-graph protocol,
following CGLB with inter-session edges preserved~\cite{zhang2022cglb}.

\paragraph{Preprocessing.}
Each graph is converted to an undirected graph, and isolated nodes are
removed. Classes are reordered by decreasing node count and then
renumbered consecutively. When Truncated SVD is applied, the projection
is fit once on the full feature matrix with a fixed random seed and is
shared by all methods and trials.

\paragraph{Class splits and sessions.}
Classes are partitioned into contiguous blocks of size $C$
(\texttt{classes\_per\_task}), yielding sessions
$\{\mathcal{T}_1,\ldots,\mathcal{T}_K\}$.
A trailing block that cannot fill $C$ classes is discarded, matching the
CGLB convention. Session $t$ introduces previously unseen classes
$\mathcal{C}_t$, and the observed class set after session $t$ is
$\mathcal{C}_{1:t}=\bigcup_{i=1}^{t}\mathcal{C}_i$.

\paragraph{Train/validation/test splits.}
Within each class, nodes are randomly split into
train/validation/test with ratios $0.6/0.2/0.2$.
The split is generated once with a fixed seed and reused across all
random trials, methods, and sessions.

\paragraph{Cumulative graphs and supervision.}
At session $t$, message passing is performed on the cumulative subgraph
induced by all nodes whose labels belong to $\mathcal{C}_{1:t}$,
including edges that connect nodes from different sessions.
Supervision uses only the current-session training nodes
$\mathcal{V}_t^{\mathrm{tr}}$; historical labels are not reused.
At test time, the model must classify among all classes in
$\mathcal{C}_{1:t}$, and the session identity of each test node is
unavailable.

\subsection{Evaluation Metrics}
\label{app:evaluation-metrics}

Let $a_{k,j}$ denote the micro accuracy on the test nodes of session
$j$ after training through session $k$ ($0\le j\le k\le K-1$).
These values form a lower-triangular accuracy matrix.

\paragraph{Per-task Micro and cumulative Micro.}
The per-task Micro entry $a_{k,j}$ is the fraction of correctly
classified test nodes of session $j$ after session $k$.
The cumulative (joint) Micro after session $k$ is the micro accuracy on
the union of test nodes from sessions $0,\ldots,k$, evaluated on the
same cumulative graph. The main tables report the final cumulative Micro
after the last session.

\paragraph{Average Performance (AP).}
Following CGLB~\cite{zhang2022cglb}, the Average Performance after
session $k$ is
\[
\mathrm{AP}_k
=
\frac{1}{k+1}
\sum_{j=0}^{k} a_{k,j}.
\]
We report the final AP, $\mathrm{AP}_{K-1}$.

\paragraph{Average Forgetting (AF).}
Average Forgetting measures the change from a session's immediate
accuracy to its accuracy after the final session:
\[
\mathrm{AF}
=
\frac{1}{K-1}
\sum_{j=0}^{K-2}
\bigl(a_{K-1,j}-a_{j,j}\bigr).
\]
Negative AF indicates forgetting; values closer to zero indicate less
forgetting (positive values indicate backward transfer).

\paragraph{Routing accuracy.}
For methods that select an expert or session-specific module without
session IDs (MAEM and TAAM), routing accuracy is the fraction of
evaluated test nodes assigned to the module that owns their true class.
After expert fusion in MAEM, ownership is defined by the fused expert's
class set.

\paragraph{Reporting protocol.}
All main results are averaged over five random seeds
$\{0,1,2,3,4\}$. We report mean $\pm$ standard deviation as percentages.
Higher Micro and AP are better; AF closer to zero is better.

\section{Detailed Baseline Descriptions}
\label{app:baselines}

Unless otherwise stated, all baselines are evaluated under the same
class-incremental expanding-graph protocol as MAEM
(Appendix~\ref{app:datasets-metrics}): training supervises only the
current-session training nodes, while message passing uses the
cumulative graph that retains inter-session edges. Classic
regularization and replay methods follow the CGLB class-IL
implementations~\cite{zhang2022cglb}; CaT, DeLoMe, PDGNNs, GSIP, and
TAAM follow their official algorithms with protocol-level adaptations
described below.

\subsection{Bare}
\label{app:baseline-bare}
Bare is sequential fine-tuning without continual-learning protection.
A shared GCN is trained on the current session only and then updated
on subsequent sessions, providing a lower bound on catastrophic
forgetting.

\subsection{EWC}
\label{app:baseline-ewc}
Elastic Weight Consolidation (EWC)~\cite{kirkpatrick2017ewc}
regularizes parameter drift with a quadratic penalty weighted by a
diagonal Fisher importance estimate computed at the end of each
session. Parameters important for previous sessions are discouraged
from changing while learning new classes.

\subsection{MAS}
\label{app:baseline-mas}
Memory Aware Synapses (MAS)~\cite{aljundi2018mas} likewise penalizes
changes to important parameters, but estimates importance online from
the sensitivity of network outputs to parameter perturbations, rather
than from the Fisher information of the classification loss. The
importance weights are accumulated across sessions.

\subsection{LwF}
\label{app:baseline-lwf}
Learning without Forgetting (LwF)~\cite{li2017lwf} preserves previous
knowledge by knowledge distillation. After each session, soft targets
from the previous model are stored; subsequent training combines the
current classification loss with a temperature-scaled distillation
loss that encourages the student to match prior predictions on the
current graph.

\subsection{TWP}
\label{app:baseline-twp}
Topology-aware Weight Preserving (TWP)~\cite{liu2021twp} extends
parameter regularization to graph structure. In addition to a
loss-based importance estimate, TWP measures the sensitivity of
topology-related edge scores and applies a stronger penalty to
parameters that are critical for preserving structural information
from earlier sessions.

\subsection{ER-GNN}
\label{app:baseline-ergnn}
Experience Replay GNN (ER-GNN)~\cite{zhou2021ergnn} maintains a
fixed-budget buffer of representative nodes selected per class
(e.g., coverage maximization or mean-of-feature sampling). When a new
session arrives, the model is trained jointly on current nodes and
buffered nodes. Following CGLB, replay graphs drop induced edges and
retain self-loops by default.

\subsection{CaT}
\label{app:baseline-cat}
Condensed Graph Memory (CaT)~\cite{liu2023cat} replaces raw node
storage with condensed synthetic features obtained by matching
distributions under multiple random GNN encoders. Each session
condenses only the isolated current-task subgraph under a class-wise
budget; later sessions replay the accumulated condensed memory
together with the current data (optionally with TIM).

\subsection{DeLoMe}
\label{app:baseline-delome}
DeLoMe~\cite{niu2024delome} performs gradient-matching condensation
with an SGC backbone. For each session it synthesizes a small set of
class-conditional features that reproduce gradients of a fresh SGC on
the current task, then trains with logit adjustment to mitigate
imbalance when replaying condensed memory.

\subsection{PDGNNs}
\label{app:baseline-pdgnn}
PDGNNs~\cite{zhang2024pdgnn} use Topology-aware Embedding Memory
(TEM): selected nodes are stored as APPNP-propagated embeddings rather
than raw features, and a decoupled APPNP+MLP classifier is trained
with replay of these topology-aware embeddings. Samplers include
random selection and coverage-based selection over APPNP receptive
fields.

\subsection{GSIP}
\label{app:baseline-gsip}
Graph Structure Information Preservation (GSIP)~\cite{li2024gsip}
builds on ER-GNN-style node replay and adds auxiliary losses that
preserve local/global structure and hidden representations of buffered
nodes. Memory selection follows coverage- or mean-feature-based
samplers; classification and structure-preservation losses are jointly
optimized on the cumulative graph.

\subsection{TAAM}
\label{app:baseline-taam}
Task-Aware Adaptive Modulation (TAAM)~\cite{liu2026taam} is a
parameter-isolation method. It freezes a shared SGC backbone and, for
each session, trains a Neural Synapse Modulator (a low-rank,
node-attentive FiLM generator) together with a session-specific
classifier head; after training, both modules are frozen. Task
identity at inference is inferred from Adaptive Multi-scale
Propagation (AMP) prototypes: APPNP-style diffusion features at
selected hops are averaged over supervised nodes and stored as a
session prototype.

The original TAAM evaluation routes an entire test subgraph to a
single session by comparing one subgraph-level AMP prototype against
stored prototypes. This assumption does not hold under our expanding
class-incremental protocol, where the cumulative graph simultaneously
contains nodes from many sessions and test evaluation provides no
session IDs. Assigning one route to the whole mixed graph would force
heterogeneous nodes through the same modulator and head.

We therefore adapt TAAM to \emph{per-node} AMP routing while keeping
the original backbone, modulator, and prototype construction.
Concretely, at session $t$ the method (i)~propagates on the full
cumulative graph, (ii)~builds the AMP prototype from only the current
session's training nodes, (iii)~trains and then freezes a new
modulator and head on those nodes, and (iv)~at inference computes an
AMP profile for every test node and assigns that node to the nearest
stored session prototype by cosine similarity before applying the
corresponding frozen modulator and head. Message passing still uses
the cumulative topology; supervision and prototype construction never
revisit previous labels. This node-level routing is necessary for
fair comparison on multi-session cumulative graphs, and is the only
protocol-level change relative to the official TAAM design.

\section{Further Experimental Results}
\label{app:further-experiments}

\subsection{Hyperparameter Analysis}
\label{app:hyperparameters}

\begin{figure}[t]
\centering
\includegraphics[width=\columnwidth]{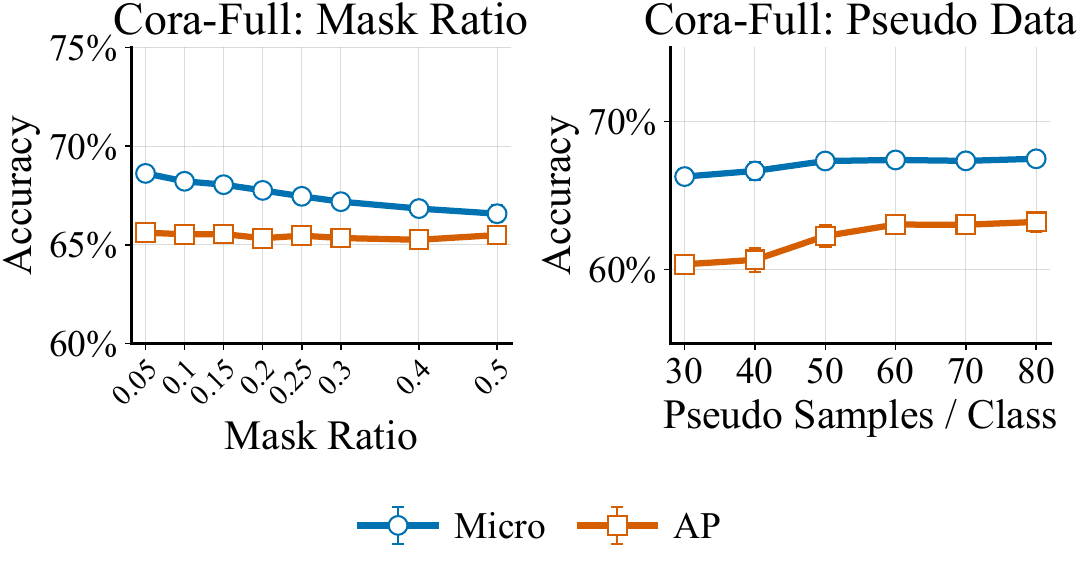}
\caption{Hyperparameter sensitivity on Cora-Full. Left: MAE mask ratio.
Right: number of pseudo samples per class used in expert fusion.
Curves report Micro and AP (mean $\pm$ std over five seeds).}
\label{fig:app-hyperparameters}
\end{figure}

We study two hyperparameters that directly affect MAE routing and
expert fusion on \textsc{Cora-Full}: the MAE mask ratio and the number
of topology-aware pseudo samples generated per class during expert
merging. All other settings follow the main experimental
configuration. Results are averaged over five seeds and are shown in
Figure~\ref{fig:app-hyperparameters}.

\paragraph{Mask ratio.}
The mask ratio controls how much of each node representation is hidden
when training and querying the session MAE. We vary it over
$\{0.05,0.1,0.15,0.2,0.25,0.3,0.4,0.5\}$.
As the mask ratio increases, Micro decreases gradually from
$68.62\%$ to $66.58\%$, while AP remains nearly stable around
$65.3\%$--$65.6\%$.
A small mask ratio preserves more observed features and yields slightly
stronger final Micro; overly aggressive masking makes reconstruction
harder and weakens the routing signal.
The default value $0.1$ used in the main experiments provides a good
balance between reconstruction difficulty and accuracy, and is close to
the best Micro in this sweep.

\paragraph{Pseudo-data sample count.}
During capacity-triggered fusion, MAEM synthesizes a fixed number of
pseudo representations per class for data-free distillation. We vary
this count over $\{30,40,50,60,70,80\}$.
Both Micro and AP improve as more pseudo samples are provided, with the
largest gains between $30$ and $60$ samples per class
(Micro from $66.28\%$ to $67.40\%$, AP from $60.36\%$ to $63.04\%$).
Beyond $60$, the curves largely plateau: increasing to $80$ yields only
marginal further improvement.
This indicates that a moderate pseudo-data budget is sufficient to
transfer knowledge into the fused expert, while a too-small budget
under-covers the teacher distributions.
We therefore use $60$ pseudo samples per class by default.

\subsection{Backbone Analysis}
\label{app:backbones}

MAEM separates graph propagation from expert learning: a fixed,
parameter-free operator first maps node features into a shared
representation $\mathbf{H}$, which is then consumed by session experts.
We compare five such operators on \textsc{ogbn-arxiv} and
\textsc{Cora-Full}, keeping all MAEM components unchanged. The hop count
is $k{=}2$ when applicable, and the PPR restart probability is
$\alpha{=}0.1$.

\begin{itemize}
    \item \textbf{Identity (Raw)}: No propagation; $\mathbf{H}=\mathbf{X}$.
    This serves as a feature-only baseline.
    \item \textbf{SGC}: Symmetric normalized propagation
    $\mathbf{H}=(\tilde{D}^{-1/2}\tilde{A}\tilde{D}^{-1/2})^k\mathbf{X}$,
    i.e., the parameter-free GCN filter used by Simplified Graph
    Convolution. This is the default backbone in the main experiments.
    \item \textbf{GraphSAGE}: Mean aggregation with equal self and
    neighbor weights,
    $h_v\leftarrow 0.5\,h_v+0.5\cdot\mathrm{mean}_{u\in\mathcal{N}(v)}h_u$,
    where the neighborhood mean excludes self-loops.
    \item \textbf{Random Walk}: Row-normalized propagation
    $\mathbf{H}=(D^{-1}A)^k\mathbf{X}$, corresponding to a random-walk
    transition matrix.
    \item \textbf{PPR}: Personalized PageRank propagation with restart,
    $\mathbf{H}^{(t+1)}=(1-\alpha)\hat{A}\mathbf{H}^{(t)}+\alpha\mathbf{X}$,
    where $\hat{A}$ is the symmetrically normalized adjacency with
    self-loops and $\alpha=0.1$.
\end{itemize}

\begin{table*}[t]
\centering
{\small
\setlength{\tabcolsep}{2pt}
\begin{tabular}{lcccccc}
\toprule
& \multicolumn{3}{c}{OGBN-Arxiv}
& \multicolumn{3}{c}{Cora-Full} \\
\cmidrule(lr){2-4}\cmidrule(lr){5-7}
Backbone
& Micro & AP & AF
& Micro & AP & AF \\
\midrule
Identity
& $33.18\pm1.65$ & $35.12\pm0.20$ & $-19.06\pm1.39$
& $43.26\pm0.25$ & $26.89\pm0.30$ & $-1.63\pm0.27$ \\
SGC / GCN
& $\mathbf{50.34\pm1.41}$ & $46.35\pm0.27$ & $\mathbf{-15.04\pm0.68}$
& $68.22\pm0.08$ & $65.54\pm0.25$ & $-7.85\pm0.36$ \\
GraphSAGE
& $43.24\pm2.03$ & $45.64\pm0.36$ & $-18.43\pm1.14$
& $\mathbf{69.72\pm0.11}$ & $65.25\pm0.25$ & $-7.04\pm0.34$ \\
Random Walk
& $48.45\pm1.60$ & $\mathbf{47.04\pm0.43}$ & $-15.75\pm1.15$
& $67.42\pm0.20$ & $\mathbf{65.86\pm0.17}$ & $-9.33\pm0.44$ \\
PPR
& $\underline{49.52\pm1.56}$ & $\underline{46.77\pm0.24}$ & $\underline{-15.63\pm0.50}$
& $\underline{69.69\pm0.17}$ & $65.16\pm0.16$ & $\mathbf{-6.02\pm0.21}$ \\
\bottomrule
\end{tabular}
}
\caption{Backbone ablation of MAEM (\%). All operators are parameter-free;
only the propagation rule changes. Mean $\pm$ std over five seeds.
Best results are in bold and second-best are underlined.}
\label{tab:app-backbones}
\end{table*}

Table~\ref{tab:app-backbones} shows that neighborhood aggregation is
consistently beneficial. On both datasets, Identity is clearly worse than
all graph-based operators, indicating that structural context is important
for MAEM's expert routing and classification.
On \textsc{ogbn-arxiv}, the default SGC backbone achieves the best
Micro ($50.34\%$) and the least forgetting among strong variants, while
Random Walk attains the highest AP ($47.04\%$).
On \textsc{Cora-Full}, GraphSAGE and PPR obtain the highest Micro
($69.72\%$ and $69.69\%$), with AP values close to SGC; Random Walk
yields the best AP ($65.86\%$), whereas Identity has the weakest AP
despite a relatively mild AF.
Overall, different parameter-free propagators remain competitive once
neighborhood information is included, and the default SGC backbone
provides a strong and stable trade-off across Micro, AP, and AF.

\subsection{Pseudo-Data Generation Quality}
\label{app:pseudo-quality}

\begin{table*}[t]
\centering
{\small
\setlength{\tabcolsep}{6pt}
\begin{tabular}{cccccc}
\toprule
Expert & Classes & $\mathrm{MMD}^{2}$
& $\mathrm{Acc}_{\mathrm{pseudo}}$
& $\mathrm{Acc}_{\mathrm{real}}$
& Ratio \\
\midrule
0 & $0$--$3$
& $0.239\pm0.019$ & $0.734\pm0.087$ & $0.903\pm0.001$ & $0.813\pm0.097$ \\
1 & $4$--$7$
& $0.170\pm0.034$ & $0.835\pm0.017$ & $0.909\pm0.002$ & $0.918\pm0.018$ \\
2 & $8$--$11$
& $0.147\pm0.029$ & $0.850\pm0.027$ & $0.955\pm0.002$ & $0.891\pm0.029$ \\
3 & $12$--$15$
& $0.187\pm0.032$ & $0.837\pm0.017$ & $0.941\pm0.003$ & $0.890\pm0.019$ \\
4 & $16$--$19$
& $0.262\pm0.060$ & $0.858\pm0.077$ & $0.932\pm0.008$ & $0.921\pm0.083$ \\
5 & $20$--$23$
& $0.162\pm0.023$ & $0.803\pm0.022$ & $0.904\pm0.011$ & $0.888\pm0.022$ \\
6 & $24$--$27$
& $0.226\pm0.021$ & $0.925\pm0.022$ & $0.976\pm0.004$ & $0.948\pm0.021$ \\
7 & $28$--$31$
& $0.270\pm0.059$ & $0.900\pm0.022$ & $0.949\pm0.007$ & $0.949\pm0.026$ \\
8 & $32$--$35$
& $0.201\pm0.021$ & $0.621\pm0.093$ & $0.845\pm0.010$ & $0.736\pm0.114$ \\
9 & $36$--$39$
& $0.328\pm0.081$ & $0.786\pm0.024$ & $0.835\pm0.024$ & $0.942\pm0.043$ \\
\midrule
Average & ---
& $0.219$ & $0.815$ & $0.915$ & $0.890$ \\
\bottomrule
\end{tabular}
}
\caption{Per-expert pseudo-embedding quality on \textsc{ogbn-arxiv}
(no-fusion, $K{=}10$). $\mathrm{MMD}^{2}$ uses a Gaussian kernel with
median-heuristic bandwidth and an unbiased
estimator.
$\mathrm{Ratio}=\mathrm{Acc}_{\mathrm{pseudo}}/\mathrm{Acc}_{\mathrm{real}}$
under an identical classifier. Mean $\pm$ std over five seeds.}
\label{tab:app-pseudo-quality}
\end{table*}

Expert fusion in MAEM relies on topology-aware pseudo embeddings
synthesized from frozen MAE experts, without revisiting historical
nodes. To assess whether these pseudo embeddings remain faithful to
the corresponding real representations, we evaluate each expert on
\textsc{ogbn-arxiv} under a no-fusion setting with $K{=}10$ experts
(four classes per expert). For every expert, we compare the
pseudo-embedding set with the real SGC embeddings of its classes.

We report three complementary quantities.
$\mathrm{MMD}^{2}$ is the squared maximum mean discrepancy between
pseudo and real embeddings, computed with a Gaussian kernel, a
median-heuristic bandwidth, and the unbiased
estimator; smaller values indicate closer
distribution matching.
$\mathrm{Acc}_{\mathrm{pseudo}}$ and $\mathrm{Acc}_{\mathrm{real}}$
are classification accuracies of the same frozen expert classifier on
pseudo and real embeddings, respectively.
Their ratio,
$\mathrm{Ratio}=\mathrm{Acc}_{\mathrm{pseudo}}/\mathrm{Acc}_{\mathrm{real}}$,
measures how much of the real decision quality is retained on
pseudo data.
Table~\ref{tab:app-pseudo-quality} summarizes the results over five
seeds.

Across experts, the average $\mathrm{MMD}^{2}$ is $0.219$, showing that
pseudo embeddings approximate but do not perfectly reproduce the real
feature distribution.
Nevertheless, the average Ratio reaches $0.890$: pseudo embeddings
recover about $89\%$ of the real-embedding accuracy
($0.815$ vs.\ $0.915$).
Several experts are especially strong (e.g., Experts $6$ and $7$ with
Ratio near $0.95$), while Expert $8$ is the weakest
($\mathrm{Acc}_{\mathrm{pseudo}}=0.621$, Ratio $=0.736$), indicating
that generation quality can vary with class structure.
Overall, the synthesized embeddings remain sufficiently class-discriminative
to support data-free distillation during expert fusion, which is consistent
with the competitive performance of capacity-controlled MAEM in the main
experiments.

\section{Limitations}
\label{app:limitations}

\paragraph{Limited expressiveness of the SGC backbone.}
MAEM uses parameter-free Simplified Graph Convolution to separate
graph propagation from expert learning. This design reduces
session-dependent feature drift and keeps the shared representation
stable across incremental sessions. However, SGC applies a fixed
linear filter without learned nonlinear message passing.
On graphs that require highly non-linear decision boundaries, long-range
relational reasoning, or heterophilic neighborhood aggregation, the
resulting representations may be insufficiently discriminative.
In such cases, expert routing and classification operate on a
constrained feature space, which can limit the attainable accuracy even
when the MoE components themselves are well trained.
Richer parameter-free or lightly parameterized backbones may alleviate
this issue, but they also reintroduce the risk of representation drift
under continual updates.

\paragraph{Sensitivity of pseudo-data synthesis to noisy SGC features.}
Expert fusion depends on topology-aware pseudo embeddings synthesized
from frozen MAE experts in the SGC feature space.
When the backbone representations are noisy---for example, under weak
homophily, severe class overlap, sparse connectivity, or poorly
separated session distributions---the MAE captures a blurred target
distribution.
Consequently, DeepInversion-style optimization may generate pseudo
samples that concentrate on ambiguous modes or under-cover difficult
classes, as also suggested by the expert-wise variation in
pseudo-embedding quality
(Appendix~\ref{app:pseudo-quality}).
Noisy features therefore propagate into the fused expert through
data-free distillation, potentially amplifying fusion errors when
capacity constraints force frequent merging.
Improving the reliability of pseudo-data generation under noisy
graph representations remains an important direction for future work.

\section{Practical MAE Reconstruction Loss}
\label{app:mae-loss}

In the main text, MAE training is written with mean squared error (MSE)
for theoretical clarity. In the experiments, however, we replace MSE by
a direction-sensitive reconstruction loss. The same practical loss is
used for MAE training, reconstruction-based routing, and pseudo-data
generation.

Node representations $\mathbf{h}\in\mathbb{R}^{d}$ are produced by
SGC. On citation and related graphs, features are often derived from
bag-of-words style attributes, so semantic information resides mainly
in the \emph{direction} of $\mathbf{h}$, whereas its magnitude can be
noisy. Plain MSE is highly sensitive to magnitude mismatch and can
destabilize training. Writing
$\tilde{\mathbf{h}}=f_t(\widehat{\mathbf{h}})$ for the MAE
reconstruction, MSE admits the decomposition
\begin{align*}
\|\mathbf{h}-\tilde{\mathbf{h}}\|_2^{2}
&=
\bigl(\|\mathbf{h}\|_2-\|\tilde{\mathbf{h}}\|_2\bigr)^{2}
+
2\|\mathbf{h}\|_2\|\tilde{\mathbf{h}}\|_2
\bigl(1-\cos\theta\bigr),
\end{align*}
where
$\cos\theta
=
(\mathbf{h}^{\top}\tilde{\mathbf{h}})
/
(\|\mathbf{h}\|_2\|\tilde{\mathbf{h}}\|_2)$.
The first term penalizes magnitude discrepancy, and the second depends
on angular disagreement. Softening the magnitude contribution therefore
reduces the influence of magnitude noise while still encouraging
directional alignment.

Direction and magnitude constraints alone, however, provide limited
discriminability for routing among multiple experts. MSE can
equivalently be expanded in terms of per-dimension statistics:
\begin{align*}
\|\mathbf{h}-\tilde{\mathbf{h}}\|_2^{2}
&=
d\Bigl[
(\mu_{\mathbf{h}}-\mu_{\tilde{\mathbf{h}}})^{2}
+
(\sigma_{\mathbf{h}}-\sigma_{\tilde{\mathbf{h}}})^{2}
+
2\sigma_{\mathbf{h}}\sigma_{\tilde{\mathbf{h}}}(1-\rho)
\Bigr],
\end{align*}
where $\mu$ and $\sigma$ denote the mean and standard deviation
across feature dimensions, $d$ is the feature dimensionality, and
$\rho$ is the Pearson correlation between $\mathbf{h}$ and
$\tilde{\mathbf{h}}$.
This motivates an explicit Pearson term that strengthens
dimension-wise structural agreement beyond cosine similarity.

Accordingly, the practical reconstruction loss used in our experiments
is
\begin{align*}
\mathcal{L}_{\mathrm{rec}}
&=
\mathcal{L}_{\mathrm{cos}}^{\gamma}
+
\alpha\,\mathcal{L}_{\mathrm{Pearson}}
+
\beta\,\mathcal{L}_{\mathrm{Norm}},
\end{align*}
where
\begin{gather*}
\mathcal{L}_{\mathrm{cos}}
=
1-
\frac{\mathbf{h}^{\top}\tilde{\mathbf{h}}}
{\|\mathbf{h}\|_2\|\tilde{\mathbf{h}}\|_2},
\\
\mathcal{L}_{\mathrm{Pearson}}
=
1-
\frac
{(\mathbf{h}-\bar{\mathbf{h}})^{\top}
(\tilde{\mathbf{h}}-\bar{\tilde{\mathbf{h}}})}
{\|\mathbf{h}-\bar{\mathbf{h}}\|_2
\|\tilde{\mathbf{h}}-\bar{\tilde{\mathbf{h}}}\|_2},
\\
\mathcal{L}_{\mathrm{Norm}}
=
\Bigl(
\log(\|\mathbf{h}\|_2+\epsilon)
-
\log(\|\tilde{\mathbf{h}}\|_2+\epsilon)
\Bigr)^{2}.
\end{gather*}
Here $\bar{\mathbf{h}}$ and $\bar{\tilde{\mathbf{h}}}$ are the
coordinate-wise means of $\mathbf{h}$ and $\tilde{\mathbf{h}}$,
$\gamma>0$ sharpens the cosine term, $\alpha,\beta\ge 0$ balance
the Pearson and magnitude regularizers, and $\epsilon>0$ ensures
numerical stability.
Unless otherwise stated, we use $\gamma=2$, $\alpha=0.6$, and
$\beta=0.3$.

Exact reconstruction $\tilde{\mathbf{h}}=\mathbf{h}$ remains a
pointwise optimum of this practical loss. Relative to MSE, it
down-weights brittle magnitude noise, preserves directional and
correlation structure that is useful for expert separation, and yields
more stable training and routing in the SGC feature space.

\end{document}